\documentclass[runningheads]{llncs}
\usepackage[T1]{fontenc}

\usepackage{graphicx}%
\usepackage{multirow}%
\usepackage{mathrsfs}%
\usepackage{xcolor}%
\usepackage{textcomp}%
\usepackage{manyfoot}%
 \usepackage{listings,url}%
\usepackage{epsfig}
\usepackage{subcaption}
\usepackage{calc}
\usepackage{amssymb}
\usepackage{amstext}
\usepackage{amsmath}
\usepackage{multicol}
\usepackage{pslatex}
\usepackage{algorithm2e}

\newcommand{\eq}[1]{\begin{align}#1\end{align}}
\newcommand{\eqs}[1]{\begin{align}#1\end{align}}
\begin{document}

%\title{A Survey of Deep Learning: From Activations to Transformers}%Performance-Focused 

\title{What comes after transformers? -- A selective survey connecting ideas in deep learning\thanks{This is an extended version of the published paper by Johannes Schneider and Michalis Vlachos titled ``A survey of deep learning: From activations to transformers'' which appeared at the International Conference on Agents and Artificial Intelligence(ICAART) in 2024. It was selected for post-publication.}}

\author{Johannes Schneider\inst{1}}
\authorrunning{J. Schneider}
\institute{University of Liechtenstein, Vaduz, Liechtenstein\\
\email{johannes.schneider@uni.li}}

% \author{\authorname{Johannes Schneider\sup{1} }
% \affiliation{\sup{1} University of Liechtenstein, Vaduz, Liechtenstein}
% \email{johannes.schneider@uni.li}
% }

% \author*[1]{\fnm{Johannes} \sur{Schneider}}\email{johannes.schneider@uni.li}
% \author[2]{\fnm{Michalis} \sur{Vlachos}}\email{     michalis.vlachos@unil.ch}
% \affil[1]{\orgname{University of Liechtenstein}, \country{Liechtenstein}}
% \affil[2]{HEC, University of Lausanne, Switzerland}
% \author{
%     Johannes Schneider$^1$ and Michalis Vlachos$^2$\\
%     $^1$ University of Liechtenstein, Liechtenstein, johannes.schneider@uni.li\\
%     $^2$ HEC, University of Lausanne, Switzerland, michalis.vlachos@unil.ch    
% }

% \author{\authorname{First Author Name\sup{1}\orcidAuthor{0000-0000-0000-0000}, Second Author Name\sup{1}\orcidAuthor{0000-0000-0000-0000} and Third Author Name\sup{2}\orcidAuthor{0000-0000-0000-0000}}
% \affiliation{\sup{1}Institute of Problem Solving, XYZ University, My Street, MyTown, MyCountry}
% \affiliation{\sup{2}Department of Computing, Main University, MySecondTown, MyCountry}
% \email{\{f\_author, s\_author\}@ips.xyz.edu, t\_author@dc.mu.edu}
% }

\maketitle 

%They include a myriad of variants related to attention, normalization, skip connections, transformers and self-supervised learning schemes -- to name a few.%While many of them might be deemed incremental, others are more ``disruptive'' in terms of their ideas but often do not live up to state-of-the-art performance (yet).
\begin{abstract}Transformers have become the de-facto standard model in artificial intelligence since 2017 despite numerous shortcomings ranging from energy inefficiency to hallucinations. Research has made a lot of progress in improving elements of transformers, and, more generally, deep learning manifesting in many proposals for architectures, layers, optimization objectives, and optimization techniques. For researchers it is difficult to keep track of such developments on a broader level. 
We provide a comprehensive overview of the many important, recent works in these areas to those who already have a basic understanding of deep learning. Our focus differs from other works, as we target specifically novel, alternative potentially disruptive approaches to transformers as well as successful ideas of recent deep learning.
We hope that such a holistic and unified treatment of influential, recent works and novel ideas helps researchers to form new connections between diverse areas of deep learning. We identify and discuss multiple patterns that summarize the key strategies for successful innovations over the last decade as well as works that can be seen as rising stars. Especially, we discuss attempts on how to improve on transformers covering (partially) proven methods such as state space models but also including far-out ideas in deep learning that seem promising despite not achieving state-of-the-art results. We also cover a discussion on recent state-of-the-art models such as OpenAI's GPT series and Meta's LLama models and, Google's Gemini model family.

%\end{keywords}
\end{abstract}
%\maketitle
%\onecolumn \maketitle \normalsize \setcounter{footnote}{0} \vfill

\section{Introduction} 
Transformers are widely regarded as the driving force behind artificial intelligence, e.g., of so-called foundation models\cite{sch24fou} such as OpenAI's ChatGPT or early models such as BERT. Transformers are at the top of most machine learning benchmark leaderboard including computer vision, speech, and natural language processing. A major advantage of deep learning is its layered, modular structure. It enables construction of models from individual components in a flexible manner. Researchers have created a large selection of layers, architectures, and objectives. Keeping up with the rapid developments in AI models is a difficult task. 

There exist multiple reviews with a narrow focus such as large language models (e.g. \cite{min21}) and convolutional neural networks (e.g. \cite{kha20}). Previous studies \cite{alom19,shre19,dong21,alz21} with a wider focus have become dated and miss new developments such as transformers and self-supervised learning. Furthermore, no survey or novel text book such as \cite{bis23} has focused on alternatives towards transformers. 
Multiple works have discussed particular shortcomings of transformers such as computational inefficiency, e.g., due to quadratic time complexity of (naive) self-attention \cite{de22att,attvis22,bra21}. Thus, there is no work that looks at recent deep learning holistically integrating general ideas in deep learning and transformers, emphasizing alternatives or ways ahead. However, taking a broader and more holistic look at different (sub-)disciplines can be highly beneficial: For instance, NLP and computer vision have reciprocally shaped one another; CNNs first emerged in computer vision and subsequently found applications in NLP, whereas transformers originated in NLP and were later incorporated into computer vision. Consequently, breaking down the barriers between disciplines proves to be highly advantageous.

This paper is driven by the goal of examining the recent advancements in deep learning from an more encompassing perspective, rather than concentrating on a specific specialized field. We contend that such an approach is crucial, especially as significant new developments have decelerated; currently, the vast majority of models employs the ``old'' but still mostly state-of-the art transformer architecture presented already in 2017\cite{Scale17}.

Providing a comprehensive overview of the field is challenging, if not impossible, given the vast number of articles published annually and the constant expansion of pertinent topics. Our strategy is to choose influential works through (i) usage statistics, (ii) specialized surveys, and (iii) for transformer alternatives, we rely on public debates (as well as our own literature search). We also offer an invigorating discussion of shared design patterns across areas that have been successful.\footnote{This paper extends the conference paper \cite{sch24sur} by engaging more deeply in discussing transfromer alternatives among other aspects}

%We not only categorize and survey methods, we also provide a critical appraisal of methods and discuss alternatives to existing methods to better understand what is truely needed and what is need and might be foregone maybe at the expense of a more complex network, e.g. batch-normalization (IP protection of batch norm, regu, init vs batchnorm) . We place more emphasis on recent methods that have been well-received by the community.
%While some older methods have become omnipresent, such as batch-normalization, dropout, multi-head attention, other newer methods are only starting to show their full potential.  Furthermore, 

% Our contributions are as follows:\
% \begin{enumerate}
%     \item 
% \end{enumerate}

% no spiking neural networks

\begin{figure*}[h]
  \centering
\includegraphics[width=0.8\linewidth]{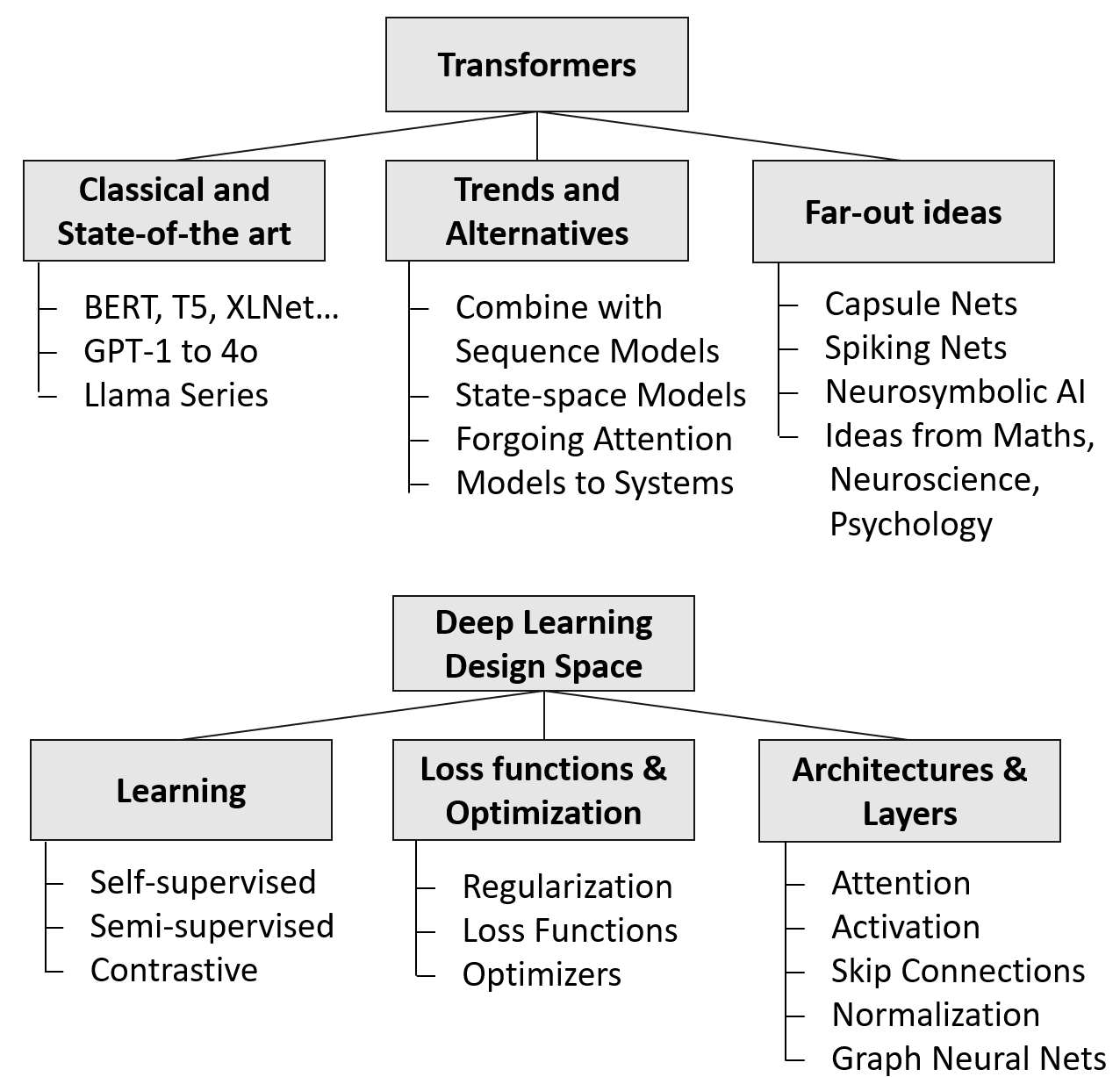}
\caption{Overview: Part 1 deals with transformers, Part 2 with deep learning \tiny{(Lower graph based on \cite{sch24sur})}} \label{fig:over}
\end{figure*}

\smallskip

\section{Overview} % of Deep Learning and Topic Selection %(for earlier works, please refer to textbooks such as \cite{goo16})%\footnote{We provide more details on the selection in the supplement. We did not cite any of our own works.}
Figure \ref{fig:over} outlines the topics addressed in this survey.
In Part 1, We start our discussion with the key topic of this survey, i.e., transformers. We refer readers with lack of knowledge of basic architectural components of transformers such as attention to Part 2 of this work or text books\cite{bis23,goo16}. However, Part 2, is also recommended for more experienced readers to reflect on recent ideas in deep learning.
In Part 1, we elaborate on current architectures including commercial and open-source models but also go beyond by discussing more far-out ideas in deep learning. The implementations of these ideas have often not yet shown state-of-the-art results or even near state-of-the-art results, but nevertheless provide interesting ideas that might just need  further thoughts. 

In Part 2, our exploration focuses on deep learning design, encompassing both objectives and training methods. Particular emphasis has been placed on work validated by usage metrics from the well-known platform "Paperswithcode.com." Our inclusion criteria feature impactful research published since 2016 and emerging contributions from 2020 onward that have quickly risen to prominence.

The coverage of each subject area is influenced by the volume of recent research and its fundamental importance. We omit discussions on data or computational strategies, such as data augmentation, model compression, and distributed machine learning, due to space constraints. This necessitated a discerning selection of model types, excluding significant categories like multi-modal models and autoencoders.

%We also had to be selective within these areas, especially, we do not discuss topics such as graph neural networks due to lack of space, and topics such as spiking neural networks, since they have not reached state-of-the-art performance. 
% This is in contrast to many prior deep learning surveys \cite{alom19,shre19,kha20} that include early history of deep learning, and while being more lengthy, miss on 

% Fine-tuning,

% maybe :working memory models, position embeddings,Parameter Norm Penalties
% Distributed Methods
% => no components specific to arch CNN

\smallskip
\section{Transformers } %(for NLP) %
Transformers have emerged as the leading architecture in deep learning. When paired with self-supervised training on extensive datasets, they have achieved top performance across numerous benchmarks in NLP (refer to \cite{liu23} for a detailed survey) and computer vision (as outlined in \cite{han2022,kha22}). Introduced in 2017 \cite{Scale17}, numerous adaptations have been developed to address challenges such as computational demands and data efficiency in the original transformer model. We discuss classical and current models in Section \ref{sec:claTra}, emerging alternatives to transformers and trends in Section \ref{sec:tre}  and more far-out ideas going beyond transformers in Section \ref{sec:far}.  %(see \cite{tay22} for a survey)

%\noindent\textbf{Transformer}: \label{sec:Trans}%\cite{Trans17}  Attention Is All You Need

It is often stated that transformers possess lower inductive bias (e.g., compared to CNNs and RNNs), making them more adaptable. Consequently, they need a larger volume of training data to offset the reduced inductive bias. Transformers are commonly pre-trained using self-supervised learning, and oftentimes fine-tuned towards specific tasks on labeled data. Classical deep learning sequence models such as recurrent neural networks (RNNs) often exhibit a time-dependency upon inference and training, e.g., after processing of one token a hidden state must be updated before the next token can be processed. This limits parallelization and leads to other issues such as vanishing gradients for long sequences. Transformers, in contrast, allow to process the entire sequence at once using self-attention to predict the next token, i.e., they are much more parallelizable than sequential models such as RNNs. Unfortunately, naive implementation of self-attention require quadratic run-time, which makes them difficult to use for long sequences.
Ideas like mixture of experts (MoE) transformer-based model have been employed to reduce computational burdens (especially during inference). MoE uses a learned routing function to select only a subset of the model for processing, which saves on computation. The idea of MoE dates back well before the year 2000 \cite{yuk12}. It has gained traction more recently as a means to reduce computational effort, in particular, during inference.

\subsection{Classical and state-of-the-art transformer architectures} \label{sec:claTra}
\begin{figure*}
%\vspace{-8pt}
  \centering
  \includegraphics[width=0.8\linewidth]{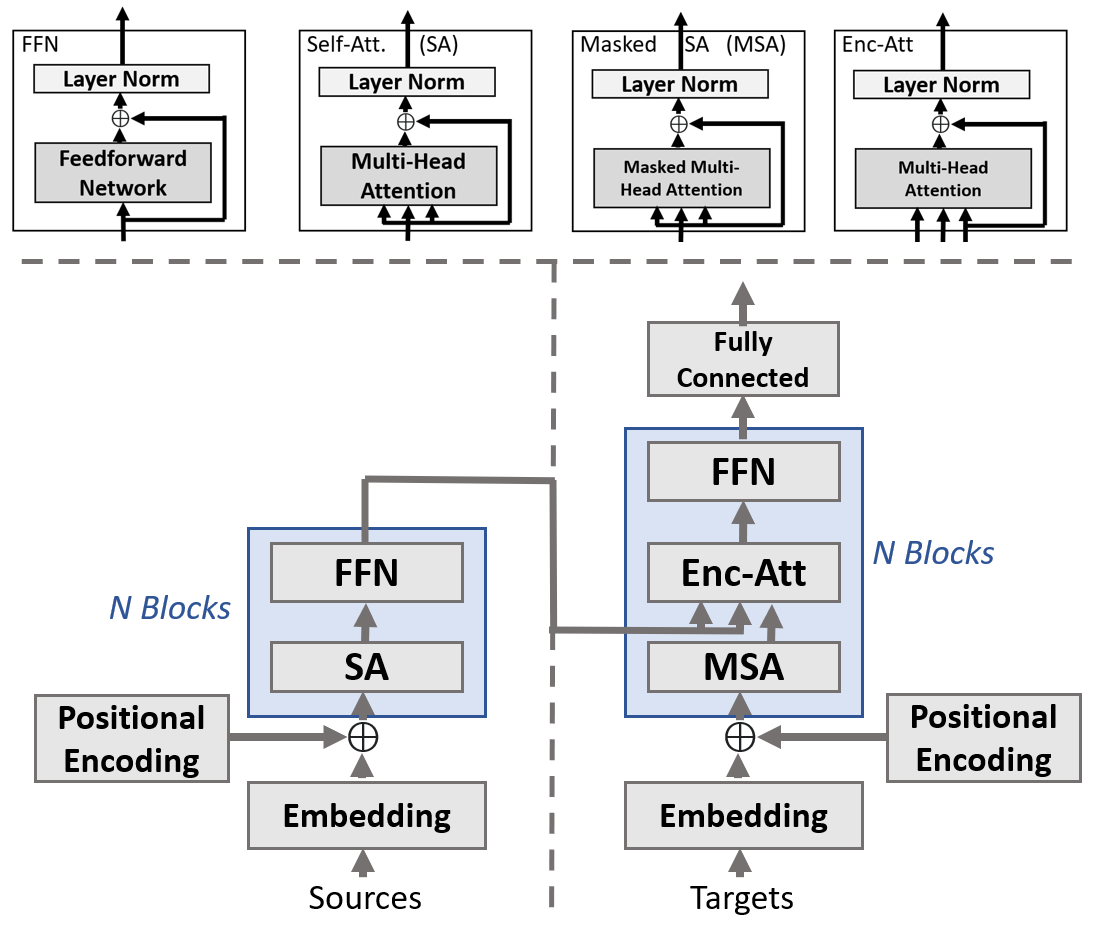}
%\vspace{-8pt}
\caption{Transformer with the four basic blocks on top and the encoder and decoder at the bottom \tiny{Figure from \cite{sch24sur}}} \label{fig:att}
%\vspace{-8pt} 
\end{figure*}

The original transformer\cite{Scale17}, designed for NLP, utilizes an encoder and decoder similar to earlier recurrent neural networks. The architecture features multiple layers of transformer blocks, as depicted in Figure \ref{fig:att}. Essential components include multi-head attention, layer normalization, and skip connections. For enhanced efficiency, several commercial and open-source models adopt the mixture of experts technique, which selectively computes outputs from a subset of models within an ensemble. Additionally, positional encodings and input embeddings are crucial. The absolute positional encodings $PE$ for position $pos$ in \cite{Scale17} are calculated using sinusoidal functions with varying frequencies:

\eq{
&\text{PE}(pos, 2i) = \sin(pos/10000^{2i/d}) \\
&\text{PE}(pos, 2i+1) =  \cos(pos/10000^{(2i)/d}) 
}
where $i$ represents the encoding dimension and $d$ the total number of dimensions. This method was chosen because it allows relative positions, potentially as significant as absolute positions, to be a linear function of the absolute position encodings.

\smallskip
\noindent\textbf{Bidirectional Encoder Representations from Transformers(BERT)} \cite{BERT18}:% \label{sec:BERT}%\cite{BERT18}  BERT: Pre-training of Deep Bidirectional Transformers for Language Understanding
The encoder of the transformer architecture generates contextual word embeddings through a masked language model pre-training objective and self-supervised learning. In this approach, the model is tasked with predicting randomly selected, masked input tokens based on their surrounding context, thus providing it with bidirectional information—that is, access to tokens both before and after the masked words. Unlike traditional next-word prediction methods, where no tokens following the predicted word are presented. Additionally, the model is required to determine whether a pair of sentences $(A,B)$ are consecutive sentences from the same document or two unrelated sentences. Once pre-trained via self-supervised methods, the model can be fine-tuned for specific tasks using labeled data.

The original BERT model has undergone various improvements, for instance, \cite{san19} minimized BERT's computational demands, and \cite{RoBER19} extended training durations, utilized longer sequences, larger batches, and more data. These enhancements have led to more robust and generalizable representations.

%\noindent\textbf{RoBERTa}: \label{sec:RoBER}%\cite{RoBER19}  RoBERTa: A Robustly Optimized BERT Pretraining Approach\cite{RoBER19} 

\smallskip
\noindent\textbf{Text-to-Text Transfer Transformer} (T5)\cite{T520} \label{sec:T5}%\cite{T519}  Exploring the Limits of Transfer Learning with a Unified Text-to-Text Transformer
considers all text-based language models as systems that produce output text from an input text. This differs from BERT\cite{BERT18} through the use of causal masking in training, which blocks the network from seeing "future" tokens of the target. Additionally, T5 diverges in its pre-training tasks.

\smallskip

\noindent\textbf{BART}\cite{BART19} is a denoising autoencoder used for pretraining sequence-to-sequence models with a conventional transformer-based machine translation architecture. It has proven effective in tasks such as language generation, translation, and comprehension. The training involves corrupting text using a variety of noising functions, including token deletion, masking, sentence permutation, and document rotation. The learning process involves reconstructing the original text from its corrupted form. The variety of noising techniques utilized can be attributed to BART's adaptation and generalization of concepts from previous models like BERT and GPT\cite{Impr18}, combining a bi-directional encoder (similar to BERT) with an autoregressive decoder (similar to GPT).

\smallskip

\noindent\textbf{XLNet} %\label{sec:XLNet}%\cite{XLNet19}  XLNet: Generalized Autoregressive Pretraining for Language Understanding %\eq{\max p(u_i|u_{i-k},...,u_{i-1})} 
\cite{XLNet19} 
This approach merges the benefits of autoregressive modeling, as seen in GPT, where it predicts the next token, with the denoising auto-encoding capabilities of BERT\cite{BERT18}, which involves reconstructing $x$ from a noisy input $\hat{x}$ created by masking words in $x$. It achieves this by using a permutation language model that samples a permutation of $Z={z_0,z_1,...,z_{T-1}}$ of the sequence $(0,1,2,...,T-1)$ aiming for the objective:
\eq{\max p(u_{z_T}|u_{z_0},...,u_{z_{T-1})}}
Importantly, the inputs themselves are not actually permuted, which would be impractical and inconsistent with later fine-tuning tasks. Instead, the permutation influences the attention mask to ensure that the factorization order determined by $Z$ is preserved.

\smallskip
\noindent The \textbf{Vision Transformer} \cite{doso20} \label{sec:Visio}%\cite{Visio20}  An Image is Worth 16x16 Words: Transformers for Image Recognition at Scale
splits an images into small patches, each of which is flattened and embedded linearly along with position embeddings. These embedded vectors for each patch are then processed by a standard transformer encoder.

\smallskip
\noindent The \textbf{Swin Transformer} \cite{Swin21} \label{sec:Swin }%\cite{Swin 21}  Swin Transformer: Hierarchical Vision Transformer using Shifted Windows
constructs hierarchical feature maps in computer vision applications instead of just a single resolution feature map. Additionally, it limits the computation of self-attention to within a local window, which helps reduce the overall computation time. 
  
\smallskip

\noindent\textbf{PaLM 2:} The original PaLM model\cite{PaLM22}, containing 540 billion parameters, parallels other prominent models like GPT-3 in scale. Its technical advancements primarily focus on the scalability of model training, enabling a single model to be trained efficiently across tens of thousands of accelerator chips. Slight modifications were made to the original transformer architecture\cite{Scale17}, such as incorporating SwiGLU activations, represented by \eq{Swish(xW)\cdot xV }, where Swish is defined in Eq. \ref{eq:swi}. Additional changes include alternative positional embeddings optimized for longer sequences, multi-query attention for faster computation, the elimination of biases for improved training stability, and shared input-output embeddings.

PaLM 2\cite{PaLM23}, the more advanced successor of PaLM, distinguishes itself by utilizing a different mixture of datasets that incorporate a broader range of languages and domains, such as programming languages and mathematics. Although it uses the traditional transformer architecture, it operates with a smaller model size yet employs more computational resources for training. Moreover, it broadens its pre-training objectives beyond simple next-word or masked-word prediction, incorporating a variety of tasks to enhance its performance.

\smallskip

\noindent\textbf{OpenAI's GPT to GPT-3 on to ChatGPT 3.5 and GPT-4o}: \label{sec:GPT-2}%\cite{GPT-219}  Language Models are Unsupervised Multitask Learners
The first version of Generative Pre-trained Transformer (GPT) is based only on the transformer's decoder to predict tokens one after the other. GPT\cite{Impr18} initially undergoes unsupervised pre-training, typically followed by supervised fine-tuning. The pre-training process involves a large corpus of tokens $U=(u_0,u_1,...,u_{n-1})$ and focuses on maximizing the probability of predicting subsequent tokens based on prior ones:
\eq{L(U)=\sum_i \log p(u_i|u_{i-k},...,u_{i-1})} where $k$ represents the context window size, and the conditional probability is computed using a neural network, specifically a multi-layer transformer decoder\cite{liu18gen} that omits the encoder as per \cite{Scale17}. Additionally, this method reduces the memory usage of the attention mechanism.\\
Building on the foundational GPT model, GPT-2 \cite{GPT-219} introduces a few key enhancements, such as relocating layer normalization to the beginning of each sub-block and adding an additional normalization after the last self-attention block. Moreover, the initialization of residual weights has been adjusted, and there has been an increase in the vocabulary, context window, and batch sizes.\\
GPT-3\cite{GPT-320} maintains a structure nearly identical to GPT-2, but it expands dramatically in scale, with more than 100 times the number of parameters and variations in the volume of training data.\\
(Chat)GPT-3.5 \cite{Chat22} is closely related to InstructGPT\cite{ouy22}, designed with a focus on adhering to user directives. InstructGPT refines GPT-3's capabilities through a dual-phase fine-tuning approach: (i) guided by demonstrations from labelers using supervised learning and (ii) leveraging human evaluations of model responses in a reinforcement learning framework. ChatGPT adheres to this methodology, where (i) supervised learning involves human AI trainers engaging in scripted dialogues, where humans act as both the user and the AI agent. These dialogues are subsequently merged with the InstructGPT conversations, which have been reformatted into dialogues. In Phase (ii), AI trainers assess the quality of responses generated by ChatGPT, choosing from multiple possibilities for a randomly chosen model-generated text. This ranking informs the reinforcement learning phase to refine the model’s performance. \\%https://openai.com/blog/chatgpt/
The technical specifications of GPT-4, the follow-up to ChatGPT-3.5, have not been officially released as per the technical report\cite{gpt423}, although some unofficial sources \cite{kdn23} claim to have insights. The report suggests that GPT-4 retains many characteristics of ChatGPT 3.5 but extends its capabilities to multi-modal functions, allowing it to process images as well. It highlights significant enhancements in training efficiency and a notable achievement in predicting the performance of large-scale models based on the outcomes of smaller models, which may have been trained on less data. This aspect is particularly crucial given the significant impact of computational costs and time on the development of extensive deep learning models.\\
Later models by OpenAI such as GPT-4 Turbo and GPT-4o \cite{gpt4o} focused on improving efficiency and response times and multi-modality, including audio, with comparable performance or modest gains on benchmarks. While there are no technical details available, models were likely down-sized and the amount of training was increased, possibly with new or improved data.

\smallskip

\noindent\textbf{Meta's Open source LLama series:} LLama is Meta's open-source model family of LLMs, ranging from LLama\cite{llama1}, LLama-2\cite{llama2} on to LLama-3.1\cite{llama3}. The LLama-3 paper also discussed multi-modal experiments. The largest model LLama 3.1 405B achieves similar performance as commercial models such as GPT4o and Claude Sonnet 3.5\cite{llama3}. All models are dense transformers rather than MoE to improve training stability. Improvements, in particular for LLama-3 originated from scaling data, models and training and changes in post-processing, e.g, using direct preference optimization (DPO)\cite{raf24} rather than reinforcement learning. That is, LLama-3 included no major architectural changes compared to prior LLama models and, in turn, to the original transformer architecture or self-supervised pre-training proceducre. Minor updates were the usage of grouped query attention \cite{ain23} and preventing self-attention between different documents (using an attention mask). Architectural details including hyperparameters are provided in \cite{llama3}.\\

%\noindent\textbf{Other competitor to GPT-series: Llama, Gemini, Mixtral, Claude-3}
\smallskip

\noindent\textbf{Gemini, Claude and other models:}
The Gemini family has been briefly described in a technical report \cite{gem23} and the version 1.5 family in \cite{gem24}. However, technical details are mostly missing. It is also based on the transformer architecture by Vaswani et al. with improvements in architecture such as multi-query attention\cite{shaz19} and optimization to help training at scale and improve efficiency. It is multi-modal including video. The version 1.5 improved long-context understanding of inputs and it introduced a sparse mixture of expert (MoE) transformer-based model. The company `Mistral.AI' also launched Mixtral, an opensource MoE model\cite{jiamix24}. GPT-4 is also alleged to be based on MoE with eight experts \cite{kdn23}. Though Anthropic's Claude models are claimed to perform well, there was no official information on its architecture or training but only a model card \cite{ant23}.

% \subsection{Vision Models}

% \noindent\textbf{EfficientNet}: \label{sec:Effic}%\cite{Effic19}  EfficientNet: Rethinking Model Scaling for Convolutional Neural Networks
% \cite{Effic19} 

% \noindent\textbf{CSPDarknet53}: \label{sec:CSPDa}%\cite{CSPDa20}  YOLOv4: Optimal Speed and Accuracy of Object Detection
% \cite{CSPDa20} 

\smallskip

\subsection{Trends and Alternatives to Transformers} \label{sec:tre}
The original transformer architecture is still the key model for LLMs and a key driver for generative AI -- as elaborated in the prior section \ref{sec:claTra}. The prior section also highlighted a few trends that emerged in the last few years to improve transformers. Many focus on computational aspects by altering existing transformers, e.g., using a MoE to reduce computation or altering attention mechanisms, while others are not driven by computation in particular and suggest more radical ideas, e.g, by combining older models like RNNs with new elements of transformers. 

\smallskip

\noindent \textbf{Combining Sequence Models and Transformers}:
That is, there is a stream of research that aims to take elements from sequence models such as RNNs and LSTMs to transformers. However, one alleged success factor for transformers in the first place is their parallelizability in contrast to classical sequence models.

\cite{bec24} proposed xLSTM, which improves on the classical LSTM cell by using exponential gating with memory mixing and a new memory structure. It allows for improved parallel training. While the results are promising, they stem mostly from smaller models (1.3B) and, thus, need to be validated using larger models.

Receptance weighted key value (RWKV) \cite{pen23} combines the well parallelizable training of transformers with the fast inference of RNNs. For inference they achieve linear time complexity in sequence length $S$ and memory needs linear in the (embedding) dimension. In contrast, classical transformer require a factor $S$ more for both time and space(memory). The architecture bares similarity with an ordinary transformer but they use a variant of linear attention to focus on channels rather than the classical dot-product token attention.

Retentive networks\cite{sun23ret} replace multi-head attention with multi-scale retention. They derive it from a recurrent model by adding context awareness. In a recurrent setting, a state $s$ is updated using a matrix $Q$. Context awareness means that the matrix $Q$ is also mulitplied by the context $X$ derived from the current input. They also perform matrix diagonalization to obtain an easily parallizable formulation.

\smallskip
\noindent \textbf{State-space models -  Replacing attention with a merger of CNNs and RNNs}:

State-space models (SSM) are well known in natural science dating back more 60 years\cite{kal60}. They can be written as $x'(t)=Ax(t)+Bu(t)$, $y(t)=Cx(t)+Du(t)$ with matrixes $A$, $B$, $C$, and $D$. They have been adopted for sequence modeling\cite{gu21co}, where they can be interpreted as ``a combination of recurrent neural networks (RNNs) and convolutional neural networks (CNNs), with inspiration from classical state space models''\cite{gu23ma}.  In particular, \cite{gu21ef} focuses on long range dependencies leading to the so-called structured state space sequence model (S4) \cite{gu21ef}. Technically, they apply a low-rank correction to matrix $A$ allowing the matrix to be diagonalized stably. Special SSM layers have also been proposed for LLMs achieving comparable performance on small models \cite{fu2022}. A major evolution is MAMBA\cite{gu23ma}, which neither requires attention nor MLPs. It introduces a selection mechanism that allows to select data in an input-dependent manner. They also create a hardware aware algorithm and simplify prior SSM architectures by merging with the MLP layer into a single block.

\smallskip
\noindent \textbf{Forgoing attention based on MLPs only and gated convolutions}:
One of the simplest architectures relies on multi-layer perceptrons (MLP)s only. The MLP-mixer\cite{tol21} has been shown to achieve state-of-the-art results in computer vision without convolutions and self-attention. It relies on MLPs that are applied on spatial locations or feature channels instead. It also maintaining other common architectural elements such as skip-connections and layer normalization. Recent work has also improved on MLPs for mixing, e.g., by using Monarch matrices \cite{fu24mon}.

A drop-in replacement for attention is Hyena \cite{pol23}. It is based on long convolutions and data-controlled gating. The paper also discusses three factors that must be fulfilled by attention alternatives but are often missed, i.e., (i) data control: Attention is a linear data controlled operator, (ii) sublinear parameter scaling: parameters (of attention layers) does not depend on sequence length, and (iii) unrestricted context: interaction between any tokens is possible, e.g., there is no locality.

\smallskip
\noindent \textbf{Models to systems - Integrating planning and external tools}:
While a single LLM is often seen or believed to be capable of doing any task, there are a number of approaches that aim at building systems rather than improving only the deep learning model in isolation. That is, LLMs are enhanced with other components that interact to improve its performance. For example, \cite{kam24} claim that large language models (LLMs) cannot plan but can generate ideas that are critiqued by separate models, which may be domain-specific and not LLMs. Ideas like Retrieval Augmented Generation (RAG)\cite{lew20} have altered the way how LLMs process prompts by first enhancing them with external knowledge, which helped to reduce hallucinations. The usage of external tools by the LLM such as calculators has also improved the performance of LLM-based systems for many tasks\cite{schi24}.

\subsection{Far-out models and ideas} \label{sec:far}
The prior section contained some interesting ideas that have arleady shown promising outcomes compared to transformers in empirical evaluations. However, one might also seek to leverage more revolutionary ideas that have been proposed in the context of deep learning but have not yet yielded the desired success.

\noindent \textbf{Capsule networks - Returning vectors instead of scalars}:
Capsule networks have been proposed in 2011\cite{hin11}. The essential idea is that a neuron outputs a vector rather than a scalar. Hinton et al. motivate the usage of a vector output in the context of computer vision. They argue that a (visual) feature is often characterized by multiple instantation parameters such as position, scale, lighting, etc. that resemble a vector rather than just a scalar. A capsule learns an implicitly defined visual entity across some limited transformations such as lighting or scaling. It returns the probability that the entity is present and a set of instantation parameters (e.g., lighting or scaling). They argue that a key advantage of this approach is the recognition of whole-part relationships, e.g., to detect a face one might identify a mouth and nose and both must have appropriate scaling and orientation. The training of the network requires pairs of transformed images to learn capsules that can extract pose parameters from inputs. Capsules themselves might consist of classical elements such as dense layers and gates. Capsule networks have been further improved since their introcduction in 2011\cite{sab17,hin18}. They managed to achieve good performance on simple daasets such as MNIST but failed to deliver close to state-of-the-art performance on challenging datasets such as ImageNet.

\smallskip
\noindent \textbf{Spiking neural networks - Adding temporality through non-synchronous activating neurons}:
Spiking neural networks \cite{maa97} exist for more than 25 years, achieving remarkable, but no state-of-the-art results. Spiking neural networks add a temporal dimension to neural networks. A neuron activates once its ``potential'' exceeds a threshold through transmission of spiking signals to its adjacent neurons. The spike increases their potential. Potentials decrease upon sending spikes but also spontaneously over time (without transmission). Thus, in contrast to non-spiking networks, transmission of signals in the network can occur in non-periodic, non-synchronized cycles among neurons. As such the time, when a signal occurs also resembles information in addition to the actual information in the spike. This effectively allows to perform time-coding as common distributed systems \cite{sch11}. 

\smallskip
\noindent \textbf{Neurosymbolic AI - Leveraging domain specific (symbolic) languages}:
There is also an (ongoing) debate on the value of symbolic approaches in conjunction with neural networks relying on continuous representations. For example, in 2017 a neuro symbolic approach for program synthesis was proposed\cite{par16} that can generate program code in a domain specific language from input-output examples. While the synthesis appears trivial compared to current program skills of LLMs, the paper still contains a number of interesting ideas, e.g., related to how to search the space of possible programs and how to incrementally refine programs. Other neuro-symbolic approaches have also performed concept extraction based on neural networks and used domain specific languages to perform quasi-symbolic program execution \cite{mao19}. A comprehensive overview with a focus on NLP is given in \cite{hami22}.

\smallskip
\noindent \textbf{Embrassing ideas from the mathematics community}:
Classical ideas from mathematics have also entered the domain of deep learning. For example, residual neural networks (ResNets) \cite{he16} essentially solve ordinary differential equations(ODE) \cite{chen18}. A simple ODE is $\frac{dy}{dt}=f(y,t)$ and the inital condition $y(t_0)=y_0$. Its solution $y_n$ at discretized time steps can be computed (recursively) using the Euler method: $$y_{n+1}=y_{n}+f(y_n,t_n)\cdot (t_{n+1}-t_n)$$
A layer $n$ in a residual network essentially performs the same computation:
$y_{n+1}=y_{n}+f(y_n,t_n)$
Thus, a sequence of residual blocks can be interpreted as a solution of the ODE with the Euler method with the initial condition $y(0)=X$, where $X$ is the input to the network. Viewing ResNets as ODEs allows to utilize mathematical tools such as ODE solvers and, in turn, might also have benefits with respect to memory and parameter efficiency. The idea of reformulating learning systems has also led to performance improvements \cite{has21}. That is, e.g., \cite{has21} introduce time-continuous RNNs described as a linear first order system.

\smallskip
\noindent \textbf{Embrassing ideas from neuroscience and psychology}:
While AI does not seek to rebuild the brain, (articial) neural networks structurally still share similarity with the real neural networks in our heads. Thus, almost any finding in the neuroscience community might be used as an inspiration for improving artificial neural networks. The main hinderness to adopting neuroscience in deep learning research is ``that we simply do not have enough information about the brain to use it as a guide''\cite{goo16}. For instance, the bain has (neurological) feedback loops: A neuron receives and processes a signal, and can then send a response back to the original source. While numerous attempts have been made to leverage this idea, overall success has been limited, i.e., it is essentially a call for more research. For example, \cite{kar19} argued that feedback (or recurrent circuits) are critical for fast object identification. In another idea from psychology, a reflective architecture was introduced that leverages explanations computed from gradients (GRADCAM) during learning but also during inference\cite{sch23} drawing inspirations from Kahneman's fast thinking (System 1) and slow (reflective) thinking (System 2).  The idea of ordering samples by difficulty to improve learning of deep learning models, i.e., curriculuum learning \cite{ben09}, can be traced back to the idea of shaping in animal training (e.g., \cite{ski58}).

% \smallskip
% \noindent \textbf{Embrassing ideas from neuroscience}
% TODO maybe for Luca (e.g. cite reflective network - CoT)

\smallskip

\section{Loss functions and Optimization}
We elaborate on fundamental loss functions and optimizers.

\smallskip

\subsection{Loss Functions}
Loss functions, as reviewed in \cite{wan20}, typically include several terms including a regularization term. While these functions are generally tailored to specific tasks, certain universal principles can be applied across various tasks. It is common to combine multiple loss terms using a weighted approach. Frequently, enhancements to previous studies are achieved merely by modifying the loss function.
 
%A survey of loss functions is given in\cite{wan20}.  There also exist surveys for specific areas, e.g., for image segmentation\cite{jad20}.
\smallskip
\noindent The \textbf{Triplet Loss} \label{sec:Tripl}%\cite{Tripl18}  Triplet Loss in Siamese Network for Object Tracking
\cite{Tripl18} was introduced for Siamese networkss, though its roots trace back to earlier work\cite{schu03}. The overarching principle involves comparing a given input with both a positive and a negative input, aiming to maximize the association with positively related inputs and minimize it with negative ones. It operates on input pairs $(x,y)$, each processed by a distinct but architecturally identical network.  The goal is to maximize the joint probability $p(x,y)$ for all pairs $(x,y)$:
\eq{L(\mathcal{V}_{p},\mathcal{V}_{n}) &=-\frac{1}{|\mathcal{V}_{p}|\cdot |\mathcal{V}_{n}|} \sum_{x \in \mathcal{V}_{p}}\sum_{y \in \mathcal{V}_{n}} \log p(x,y)  \\
&=-\frac{1}{|\mathcal{V}_{p}|\cdot |\mathcal{V}_{n}|} \sum_{x \in \mathcal{V}_{p}}\sum_{y \in \mathcal{V}_{n}} \log (1+e^{x-y}) } \\
\noindent Here, $\mathcal{V}_{p}$ and $\mathcal{V}_{n}$ are the positive and negative score set respectively.

% \noindent\textbf{InfoNCE}: \label{sec:InfoN}%\cite{InfoN18}  Representation Learning with Contrastive Predictive Coding
% \cite{InfoN18} 

% Too Old \noindent\textbf{GAN Least Squares Loss}: \label{sec:GAN L}%\cite{GAN L16}  Least Squares Generative Adversarial Networks
% \cite{GAN L16} 

%Too old\noindent\textbf{NT-Xent}: \label{sec:NT-Xe}%\cite{NT-Xe16}  Improved Deep Metric Learning with Multi-class N-pair Loss Objective
%\cite{NT-Xe16} 

\smallskip
\noindent The \textbf{Supervised Contrastive Loss}\cite{Super20}  \label{sec:Super}%\cite{Super20}  Supervised Contrastive Learning
groups clusters of points from the same class together in the embedding space while pushing apart samples from different classes. It is designed to utilize label information more effectively than cross-entropy loss, optimizing the separation and aggregation based on class identity.
%Rephrase next
\eqs{\mathcal{L}_i^{sup}&=\frac{-1}{2N_{\boldsymbol{\tilde{y}}_i}-1}\cdot \\
&\sum_{j=1}^{2N}\mathbf{1}_{i\neq j}\cdot\mathbf{1}_{\boldsymbol{\tilde{y}}_i=\boldsymbol{\tilde{y}}_j}\cdot\log{\frac{\exp{(\boldsymbol{z}_i\cdot\boldsymbol{z}_j/\tau)}}{\sum_{k=1}^{2N}\mathbf{1}_{i\neq k}\cdot\exp{(\boldsymbol{z}_i\cdot\boldsymbol{z}_k/\tau)}}}
} 
\noindent where $N_{\boldsymbol{\tilde{y}}_i}$ denotes the total number of images in the minibatch with the same label $\boldsymbol{\tilde{y}}_i$ as the anchor $i$. The overall loss is calculated as the sum of the losses for all anchor points  $i$, i.e., $\mathcal{L}=\sum_i\mathcal{L}_i^{sup}$. This loss formulation is particularly advantageous for supervised learning due to its:
\begin{itemize}
    \item Ability to generalize across an arbitrary number of positive examples.
    \item Increased contrastive effectiveness as the number of negative examples grows.
\end{itemize}

\smallskip
\noindent The \textbf{Cycle Consistency Loss}\cite{Cycle17} \label{sec:Cycle}%\cite{Cycle17}  Unpaired Image-to-Image Translation using Cycle-Consistent Adversarial Networks
is specifically designed for unpaired image-to-image translation using generative adversarial networks. It facilitates the learning of mappings between two distinct image domains $X$ and $Y$. Optimizing the loss supports the learning of mappings $G: X \rightarrow Y$ and $F: Y \rightarrow X$ so that one reverses the other, i.e., $F(G(x))\approx x$ and $G(F(y)) \approx y$.
\eqs{L(G,F)&= \mathbb{E}_{x \sim p_{data}(x)}[||F(G(x)) - x||_{1}]\\
&+ \mathbb{E}_{y \sim p_{data}(y)}[||G(F(y)) - y||_{1}]}
\smallskip

\smallskip
\noindent\textbf{Focal Loss}\cite{Focal17} \label{sec:Focal}%\cite{Focal17}  Focal Loss for Dense Object Detection
modifies the standard cross-entropy loss to concentrate learning efforts on hard-to-classify samples. It incorporates a factor $(1-p)^{\gamma}$, where $p$ is the probability of a sample from the cross entropy loss and $\gamma$ is a freely adjustable parameter.
\eq{L(p)=(1-p)^{\gamma}\log(p)}

\subsection{Regularization}

Regularization techniques, as reviewed in \cite{mor20}, are critically beneficial for deep learning applications. Explicit regularization incorporates an additional loss term $R(f)$ for the network $f$ into the loss function $L(x)$ for data $(x_i,y_i)$ with a balancing parameter $\lambda$.
\eq{\min_f \sum_i L(x_i,y_i)+\lambda R(f)}
Implicit regularization encompasses all other forms of regularization, such as early stopping or employing a robust loss function. Common methods like $L2$-regularization and dropout\cite{sri14}, which involves setting the activations of a random subset of neurons to zero, are among the most extensively used techniques.% and the latter exists in many variations. 

\smallskip
\noindent\textbf{Entropy Regularization} \label{sec:Entro}
\cite{Entro16} is designed to promote diversity, particularly through asynchronous methods in deep reinforcement learning \cite{wil91,Entro16}. This approach enhances the diversity of actions in reinforcement learning by avoiding excessive optimization towards only a small portion of the environment. Entropy is calculated based on the probability distribution of actions as determined by the policy as:
 \eq{H(x)=\sum_x \pi(x)\cdot\log(\pi(x))}

%REF NOT CORRECT \noindent\textbf{ALS}: \label{sec:ALS}%\cite{ALS20}  Efficient Model for Image Classification With Regularization Tricks \cite{ALS20} 

%NOt clear how relevant - rank regularization in ref, but seems not the key concept of paper\noindent\textbf{SCN}: \label{sec:SCN}%\cite{SCN20}  Suppressing Uncertainties for Large-Scale Facial Expression Recognition \cite{SCN20} 

%\cite{Path 19}  Analyzing and Improving the Image Quality of StyleGAN
\smallskip
\noindent\textbf{Path Length Regularization} \cite{Path19} \label{sec:Path }
\cite{Path19} focuses on enhancing the image quality for generative adversarial networks by ensuring consistent step lengths in the latent space and changes in the generated images. That is, changes in the latent space representation lead to (proportional) changes in the generated image (and vice versa). The principle behind this is to facilitate a fixed-size step in the latent space $\mathcal{W}$  to yield a consistent, non-zero change in image magnitude. This approach aims to improve the conditioning of GANs, which can simplify the process of architecture search and generator inversion. Gradients, with respect to $\mathbf{w} \in \mathcal{W}$, originating from random directions in the image space should be almost equal in length independent of $\mathbf{w}$ or the image space direction. The local metric scaling characteristics of the generator $g: \mathcal{W} \rightarrow \mathcal{Y}$ are stated by the Jacobian matrix $\mathbf{J_{w}} = \delta{g}(\mathbf{w})/\delta{\mathbf{w}}$. The formulation for the regularizer is:

\eq{ \mathbb{E}_{\mathbf{w},\mathbf{y} \sim \mathcal{N}(0, \mathbf{I})} (||\mathbf{J}^{\mathbf{T}}_{\mathbf{w}}\mathbf{y}||_{2} - a)^{2}
}
\noindent where $y$ corresponds to random images with pixel values that follow a normal distribution, and $w \sim f(z)$, where $z$ is also normally distributed. The constant $a$ is defined as the exponential moving average of $||\mathbf{J}^{\mathbf{T}}_{\mathbf{w}}\mathbf{y}||_{2}$. The paper also avoids the computationally expensive, explicit computation of the Jacobian.

\smallskip
\noindent\textbf{DropBlock}\cite{DropB18} \label{sec:DropB}%\cite{DropB18}  DropBlock: A regularization method for convolutional networks
 selectively drops correlated regions within feature maps instead of dropping features independently. This method is particularly effective for convolutional neural networks, where feature maps tend to show spatial correlations and a real-world feature typically corresponds to a contiguous spatial area within these maps.

\smallskip
\noindent\textbf{$R_1$ Regularization} \label{sec:R1 Re}%\cite{R1 Re18}  Which Training Methods for GANs do actually Converge? 
\cite{regr1} penalizes the  discriminator in generative adversarial networks based on the gradient to stabilize training:
\eq{
R_{1}(\psi) = \frac{\gamma}{2}E_{p_{D}(x)}[||\nabla{D_{\psi}(x)}||^{2}]}
Technically, the regularization term imposes penalties on gradients that are orthogonal to the data manifold, aiming to refine the learning process.

\smallskip

\subsection{Optimization} %,sun19
Optimization, as explored in \cite{sun20}, involves determining the best set of network parameters to minimize the loss function. The two most widely recognized techniques are stochastic gradient descent (SGD) and Adam. Neither technique consistently surpasses the other across all scenarios in terms of generalization performance. SGD has historical roots extending back to the 1950s \cite{kief52}, whereas Adam was developed more recently, in 2014 \cite{Adam14}.

\smallskip
\noindent\textbf{Adafactor} \label{sec:Adafa}%\cite{Adafa18}  Adafactor: Adaptive Learning Rates with Sublinear Memory Cost
 \cite{Adafa18} enhances the efficiency of the Adam optimization algorithm by reducing its memory requirements. This is achieved by keeping only row- and column-wise statistics of parameter matrices, instead of storing detailed per-element information.
 
\smallskip
\noindent\textbf{Layerwise adaptive large batch optimization (LAMB)}\cite{LAMB19} \label{sec:LAMB}%\cite{LAMB19}  Large Batch Optimization for Deep Learning: Training BERT in 76 minutes
builds on Adam and accelerates training by utilizing large mini-batches. It applies per-dimension and layerwise normalization to further enhance the optimization process.

%\eq{&r_{t} = \frac{m_{t}}{\sqrt{v_{t}} + \epsilon}\\
%&x_{t+1}^{(i)} = x_{t}^{(i)}  - \eta_{t}\frac{\phi(|| x_{t}^{(i)} ||)}{|| m_{t}^{(i)} || }(r_{t}^{(i)}+\lambda{x_{t}^{(i)}})}
%the adaptivity of LAMB is two-fold: (i) per dimension normalization with respect to the square root of the second moment used in Adam and (ii) layerwise normalization obtained due to layerwise adaptivity.

%Not interesting \noindent\textbf{Deep Ensembles}: \label{sec:Deep }%\cite{Deep 16}  Simple and Scalable Predictive Uncertainty Estimation using Deep Ensembles \cite{Deep 16} 

\smallskip
\noindent\textbf{Decoupled Weight Decay Regularization for ADAM}: \label{sec:AdamW}%\cite{AdamW17}  Decoupled Weight Decay Regularization
% %https://paperswithcode.com/method/adamw
% %https://towardsdatascience.com/why-adamw-matters-736223f31b5d
AdamW\cite{AdamW17} leverage a seemingly trivial observation: The original Adam optimizer updates weights with (L2-)regularization after calculating the gradients for Adam. However, it intuitively makes more sense that moving averages of gradients should exclude regularization.

%distributed \noindent\textbf{Local SGD}: \label{sec:Local}%\cite{Local18}  Local SGD Converges Fast and Communicates Little \cite{Local18} 

%\noindent\textbf{LARS}: \label{sec:LARS}%\cite{LARS17}  Large Batch Training of Convolutional Networks \cite{LARS17} 

%\noindent\textbf{AMSGrad}: \label{sec:AMSGr}%\cite{AMSGr19}  On the Convergence of Adam and Beyond\cite{AMSGr19} 
\smallskip
\noindent\textbf{RAdam and AMSGrad}: %\label{sec:RAdam}%\cite{RAdam19}  On the Variance of the Adaptive Learning Rate and Beyond
Both techniques address the convergence issues associated with Adam. Rectified Adam (RAdam)\cite{RAdam19} adjusts the variance of the adaptive learning rate, which is initially large. This approach is akin to the warm-up heuristic, where small initial learning rates are beneficial. AMSGrad\cite{AMSGr19} uses the maximum of past squared gradients instead of the exponential average to enhance stability and convergence.

\smallskip
\noindent\textbf{Stochastic Weight Averaging}: \label{sec:Stoch}%\cite{Stoch18}  Averaging Weights Leads to Wider Optima and Better Generalization
Naive averaging of weights from different epochs during stochastic gradient descent, using either a constant or cycling learning rate, has been shown to improve performance\cite{Stoch18}.

\smallskip
\noindent\textbf{Two Time-scale Update Rule(TTUR)}: \label{sec:TTUR}%\cite{TTUR17}  GANs Trained by a Two Time-Scale Update Rule Converge to a Local Nash Equilibrium
TTUR\cite{TTUR17} enhances the training of generative adversarial networks by employing separate learning rates for the discriminator and the generator. When the generator is fixed, the discriminator can reach a local minimum. This approach remains effective even if the generator converges slowly, for instance, by using a smaller learning rate. By allowing the generator to more deeply assimilate feedback from the discriminator before exploring new regions, this method aids in the convergence of GANs and can lead to improved performance.
Besides ensuring convergence, the performance may also improve since the discriminator must first learn new patterns before they are transferred to the generator. In contrast, a generator which is overly fast, drives the discriminator steadily into new regions without capturing its gathered information.

%\noindent\textbf{DFA}: \label{sec:DFA}%\cite{DFA16}  Direct Feedback Alignment Provides Learning in Deep Neural Networks \cite{DFA16} 

%could add: \noindent\textbf{KP}: \label{sec:KP}%\cite{KP19}  Deep Learning without Weight Transport \cite{KP19} 

%not main interest \noindent\textbf{Gradient Sparsification}: \label{sec:Gradi}%\cite{Gradi17}  Gradient Sparsification for Communication-Efficient Distributed Optimization \cite{Gradi17} 

%wrong \noindent\textbf{Gravity}: \label{sec:Gravi}%\cite{Gravi21}  Gravity Optimizer: a Kinematic Approach on Optimization in Deep Learning \cite{Gravi21} 

%not so good \noindent\textbf{MAS}: \label{sec:MAS}%\cite{MAS20}  Mixing ADAM and SGD: a Combined Optimization Method \cite{MAS20} 

% not so good \noindent\textbf{AMP}: \label{sec:AMP}%\cite{AMP20}  Regularizing Neural Networks via Adversarial Model Perturbation \cite{AMP20} 

\smallskip
\noindent\textbf{Sharpness-Aware Minimization}\cite{Sharp20} \label{sec:Sharp}%\cite{Sharp20}  Sharpness-Aware Minimization for Efficiently Improving Generalization
minimizes both the loss value and sharpness, which enhances generalization. It seeks parameters within neighborhoods of low loss values, rather than focusing solely on parameters that individually exhibit low loss. The loss function is defined as:
\eq{
\min_w \max_{||\epsilon||_p\leq \rho} L(w+\epsilon)
}

\smallskip

\section{Self, Semi-supervised and Contrastive learning} %(\cite{liu21self} for a survey)
Semi-supervised learning utilizes a large volume of unlabeled data alongside a small amount of labeled data, as reviewed in \cite{yang22}. Self-supervised learning, on the other hand, generates (pseudo) labels from artificial tasks, reducing the need for manually labeled data. Both approaches alleviate the burden of collecting human-annotated data. Combining self-supervised (pre-)training with fine-tuning on a small human-labeled dataset can achieve state-of-the-art results. This paradigm has significantly expanded in recent years (surveyed in \cite{eric22}). It is often integrated with contrastive learning, which aims to learn the distinction between similar and dissimilar data. By automatically distorting data to varying degrees, creating "pseudo-labeled" data for self-supervised learning becomes straightforward.

% \noindent\textbf{Inpainting}: \label{sec:Inpai}%\cite{Inpai16}  Context Encoders: Feature Learning by Inpainting
% \cite{Inpai16} 

% \noindent\textbf{MoCo}: \label{sec:MoCo}%\cite{MoCo19}  Momentum Contrast for Unsupervised Visual Representation Learning
% \cite{MoCo19} 

% \noindent\textbf{Colorization}: \label{sec:Color}%\cite{Color16}  Colorful Image Colorization
% \cite{Color16} 

% \noindent\textbf{Contrastive Predictive Coding}: \label{sec:Contr}%\cite{Contr18}  Representation Learning with Contrastive Predictive Coding
% \cite{Contr18} 

% \noindent\textbf{Jigsaw}: \label{sec:Jigsa}%\cite{Jigsa16}  Unsupervised Learning of Visual Representations by Solving Jigsaw Puzzles
% \cite{Jigsa16} 

\smallskip
\noindent The \textbf{simple framework for contrastive learning} (SimCLR)\cite{SimCL20} \label{sec:SimCL}%\cite{SimCL20}  A Simple Framework for Contrastive Learning of Visual Representations
maximizes the agreement between two inputs derived from different augmentations of the same data sample. Augmentations can include random cropping, color distortions, and Gaussian blur. A standard ResNet\cite{he16} is used to obtain representation vectors, which are then further processed through a simple MLP before applying the contrastive loss.

\smallskip
\noindent\textbf{Bootstrap Your Own Latent (BYOL)} \label{sec:BYOL}%\cite{BYOL20}  Bootstrap Your Own Latent - A New Approach to Self-Supervised Learning
\cite{BYOL20} employs both an online network and a target network, each with identical architectures comprising an encoder, a projector, and a predictor, but with separate weights. The parameters of the target network are updated as an exponential moving average of the online network's parameters. The task of the online network is to predict the representation of the target network given an augmentation of the same input.

% \noindent\textbf{DINO}: \label{sec:DINO}%\cite{DINO21}  Emerging Properties in Self-Supervised Vision Transformers
% \cite{DINO21} 

\smallskip
\noindent\textbf{Barlow Twins}\cite{barl19} \label{sec:Barlo}%\cite{Barlo21}  Barlow Twins: Self-Supervised Learning via Redundancy Reduction
rely on an objective function that aims to reduce cross-correlation $C$ between outputs for a set of image $Y^A$ and their distorted versions $Y^B$ as close to the identity as possible, i.e., the loss (including  $\lambda$ as a tuning parameter) is:

use an objective function aiming to minimize the cross-correlation 
$C$ between outputs for a set of images $Y^A$ and their distorted versions $Y^B$, bringing it as close to the identity matrix as possible. The loss function, incorporating $\lambda$ as a tuning parameter, is given by:
\eq{L=\sum_i (1-C_{i,i})^2+\lambda\cdot\sum_i\sum_{j\neq i} C_{i,j}^2}

\smallskip
\noindent\textbf{Momentum Contrast (MoCo)} \label{sec:MoCo}%\cite{MoCo19}  Momentum Contrast for Unsupervised Visual Representation Learning
\cite{MoCo19} constructs a dynamic dictionary via an encoder using unsupervised contrastive learning. During training, it performs look-ups, ensuring that an encoded query closely matches its corresponding encoded key while being dissimilar to other keys. The dictionary functions as a queue of data samples: for each mini-batch, new encoded samples are added, and the oldest mini-batch is dequeued. The key encoder's parameters are updated using a momentum-based moving average of the query encoder, promoting consistency over time.

% \noindent\textbf{Contrastive Predictive Coding}: \label{sec:Contr}%\cite{Contr18}  Representation Learning with Contrastive Predictive Coding
% \cite{Contr18} 

\smallskip
\noindent\textbf{Noisy Student}: \label{sec:Noisy}%\cite{Noisy19}  Self-training with Noisy Student improves ImageNet classification
\cite{Noisy19} outlines a method where an CNN-based EfficientNet model is initially trained on labeled data. This trained model then acts as a teacher, generating pseudo labels for unlabeled images. Subsequently, a larger model is trained on the combined labeled and pseudo-labeled data. This process is iteratively repeated, with each student model becoming the teacher for the next iteration. During the training of the student model, techniques like dropout and data augmentation are employed to introduce noise, making the learning process more challenging and allowing the student to surpass the performance of the teacher.

\smallskip
\noindent\textbf{FixMatch} \cite{FixMa20} \label{sec:FixMa}%\cite{FixMa20}  FixMatch: Simplifying Semi-Supervised Learning with Consistency and Confidence
predicts the label of a weakly-augmented image. If the confidence in this prediction exceeds a certain threshold, the model is then trained to produce the same label for a strongly-augmented version of the image.

% \noindent\textbf{MoCo v2}: \label{sec:MoCo }%\cite{MoCo 20}  Improved Baselines with Momentum Contrastive Learning

\smallskip

\section{Architectures and Layers}
We elaborate on four key types of layers: activation layers, skip connections, normalization layers, and attention layers. This is followed by a discussion of various contemporary architectures based on transformers and graph neural networks.

\smallskip

\subsection{Activation}
Activation functions are typically non-linear and significantly influence gradient flow and learning. Early activation functions, such as sigmoid and tanh, were commonly used from the 1960s through the early 2000s. However, these functions can cause training difficulties in deep networks due to the vanishing gradient problem when they saturate. The introduction of the rectified linear unit (ReLU) in 2010\cite{nai10} was a breakthrough. Although the original ReLU remains widely used, transformer architectures have introduced other activation functions and ReLU variants. Most of these alternatives share the qualitative behavior of ReLU, where outputs for negative inputs are of small magnitude and outputs for positive inputs are unbounded (see \cite{api21} for a survey).

\smallskip
\noindent\textbf{Gaussian Error Linear Units (GELU)}\cite{GELU16}  \label{sec:GELU}%\cite{GELU16}  Gaussian Error Linear Units (GELUs)
These functions weigh inputs by their percentile (ReLUs only use the sign). The activation function is the product of the input and the standard Gaussian cumulative distribution function $\Phi(x)$, i.e.,
\eq{GELU(x)= x\cdot \Phi(x)}
%\\text{GELU}(x) = x{P}(X\\leq{x}) = x\\Phi(x) = x \\cdot \\frac{1}{2}[1 + \\text{erf}(x/\\sqrt{2})],$$\r\nif $X\\sim \\mathcal{N}(0,1)$.

% \noindent\textbf{GLU}: \label{sec:GLU}%\cite{GLU16}  Language Modeling with Gated Convolutional Networks
% \cite{GLU16} 
% https://paperswithcode.com/method/glu

\smallskip
\noindent The \textbf{Mish} activation\cite{Mish19} \label{sec:Mish}%\cite{Mish19}  Mish: A Self Regularized Non-Monotonic Activation Function
originates from systematic search inspired by Swish and ReLU: 
\eq{&f(x)=x\cdot \tanh(soft^+(x))\\
&\text{ with } soft^+(x):=\ln(1+e^x)
}
In contrast, the Swish activation\cite{Swish17}  is:
\eq{&f(x)=x\cdot sigmoid(\beta x) \label{eq:swi}} Here $\beta$ is a learnable parameter.

%\noindent\textbf{Swish}: \label{sec:Swish}%\cite{Swish17}  Searching for Activation Functions

%\noindent\textbf{Hard Swish}: \label{sec:Hard }%\cite{Hard 19}  Searching for MobileNetV3 \cite{Hard 19}

\smallskip

\subsection{Skip connections}
Skip connections were introduced for residual networks\cite{he16}. In their simplest form, the output $y$ for an input $x$ of a block $L$ of layers with a skip connection is $y(x)=L(x)+x$.
The term "residual" was used in the original paper because the layer 
$L$ is tasked with learning a residua $L(x)=H(x)-x$ rather than the desired mapping $H$ itself. Since then, skip connections have been utilizied in a number of variations.

\smallskip

\noindent\textbf{ResNeXt Block}\cite{ResNe16}: \label{sec:ResNe}%\cite{ResNe16}  Aggregated Residual Transformations for Deep Neural Networks
This split-transform-merge approach for residual blocks involves evaluating a set of residual blocks concurrently and then aggregating their outputs back into one output.
\smallskip
\noindent A \textbf{Dense Block}\cite{Dense16} \label{sec:Dense}%\cite{Dense16}  Densely Connected Convolutional Networks
receives inputs from all preceding layers with matching feature map sizes and connects to all such subsequent layers.

\smallskip
\noindent\textbf{Inverted Residual Block}\cite{Inver18}: \label{sec:Inver}%\cite{Inver18}  MobileNetV2: Inverted Residuals and Linear Bottlenecks
By reversing the channel width sequence to a narrow-wide-narrow order from the original wide-narrow-wide configuration\cite{he16}, and using depthwise convolutions for the wide layer, parameters are reduced, and residual blocks execute more quickly. Furthermore, the activation function of the last layer within a block is skipped.

\smallskip

\subsection{Normalization}
With the advent of batch normalization\cite{iof15}, the concept of normalization has significantly enhanced training speed, stability, and generalization in neural networks including almost all architectures. However, its necessity is debated\cite{shao20}; for certain applications, careful initialization and learning rate adjustments may render normalization partially redundant.\\
The principle behind normalization is to transform a value $x$ to a normalized value $\tilde{x}$, by subtracting the mean $\mu$ and scaling by the standard deviation $\sigma$, i.e.,
$\tilde{x} = \frac{x-\mu}{\sigma}$.
Normalization approaches differ in the computation of $\mu$ and $\sigma$, e.g., $\mu$ and $\sigma$ can be computed across different channels. 

\smallskip
\noindent\textbf{Layer Normalization}: %\cite{Layer16}  Layer Normalization
Normalization statistics are calculated using summed inputs\cite{Layer16} for a layer $L$ with $|L|$ neurons as:
\eqs{&\mu = \frac{1}{|L|} \sum_{i=0}^{|L|-1} a_i \text{\phantom{abcd}} \sigma =  \sqrt{\frac{1}{|L|} \sum_{i=0}^{|L|-1} (a_i-\mu)^2} }
Unlike batch normalization, this method does not limit the batch size and eliminates inter-batch dependencies, making it suitable even for batch sizes as small as one.

\smallskip
\noindent\textbf{LayerScale}\cite{Layer21} %\cite{Layer21}  Going deeper with Image Transformers
has been established for transformers as a per-channel multiplication of outputs of a residual block with a diagonal matrix:
\eq{
&x_{l'}= x_{l}+diag(\lambda_1,...,\lambda_{d})\cdot SA(\eta(x))\\
&x_{l+1}= x_{l'}+diag(\lambda_1,...,\lambda_{d})\cdot FFN(\eta(x))
}
$SA$ is the self-attention layer, $FFN$ is the feed forward network, and $\eta$ the layer-normalisation (see Figure \ref{fig:att}).

\smallskip
\noindent\textbf{Instance Normalization}\cite{Insta16} \label{sec:Insta}%\cite{Insta16}  Instance Normalization: The Missing Ingredient for Fast Stylization
calculates for a four-dimensional input, such as an image of height $H$, width $W$, channels $C$, and batch size $T$:
\eq{&\mu_{t,c} = \frac{1}{HWT} \sum_{t<T,w<W,h<H} x_{t,c,w,h}\\
&\sigma_{t,c} =  \sqrt{\frac{1}{HWT} \sum_{t<T,w<W,h<H} (x_{t,c,w,h}-\mu_{t,c})^2}
}
This method can be applied, for example, to normalize image contrast. Various adaptations exist, including a version that scales according to weight norms\cite{Weigh19}.%s a commonly used adaptive version that has no learnable parameters \cite{Adapt17} as well as
%\noindent\textbf{Adaptive Instance Normalization}: \label{sec:Adapt}%\cite{Adapt17}  Arbitrary Style Transfer in Real-time with Adaptive Instance Normalization
% \noindent\textbf{Weight Demodulation}: \label{sec:Weigh}%\cite{Weigh19}  Analyzing and Improving the Image Quality of StyleGAN \cite{Weigh19} 

\smallskip

\subsection{Attention}
Attention mechanisms, explored in \cite{bra21,attvis22}, enable the learning of relevance scores for inputs, mimicking cognitive attention processes. This allows certain parts of the inputs to be highlighted as highly important, while others may be ignored as irrelevant. Often, the importance of a specific input is influenced by its context; for example, the significance of a word in a text document usually relies on the words surrounding it. %For an overview of attention in general see \cite{bra21}, and for a more recent review in computer vision see \cite{attvis22}.

\smallskip
\noindent\textbf{Scaled Dot-Product Multi-Head Attention} \cite{Scale17}: \label{sec:Scale}%\cite{Scale17}  Attention Is All You Need

The use of dot products, coupled with down-scaling, has been highly effective in calculating attention scores. %though alternatives have been proposed recently %\cite{}%https://medium.com/@madali.nabil97/do-we-really-need-the-scaled-dot-product-attention-76cbc0cb7276}
Attention takes a query $Q$, a key $K$ and a value $V$ as inputs and outputs an attention score:
\eq{\text{Att}(Q, K, V) = \text{softmax}\big(\frac{QK^{T}}{\sqrt{d_k}}\big)\cdot V}
%If we assume that $q$ and $k$ are $d_k$-dimensional vectors whose components are independent random variables with mean $0$ and variance $1$, then their dot product, $q \cdot k = \sum_{i=1}^{d_k} u_iv_i$, has mean $0$ and variance $d_k$.  Since we would prefer these values to have variance $1$, we divide by $\sqrt{d_k}$.",

%\noindent\textbf{Multi-Head Attention}: \label{sec:Multi}%\cite{Multi17}  Attention Is All You Need

Employing multiple, independent attention mechanisms in parallel enables the system to focus on different aspects of the input simultaneously. In the case of multi-head attention, matrices \textbf{W} are learned to facilitate this process. Formally:
\eq{
&\text{MultiHead}(\textbf{Q}, \textbf{K}, \textbf{V}) = [\text{h}_{0},\dots,\text{h}_{n-1}]\textbf{W}_{0}\\
&\text{where } \text{head h}_{i} = \text{Att} (\textbf{Q}\textbf{W}_{i}^{Q}, \textbf{K}\textbf{W}_{i}^{K}, \textbf{V}\textbf{W}_{i}^{V} ) }

\smallskip
\noindent\textbf{Factorized (Self-)Attention} \cite{Fixed19} \label{sec:Fixed}%\cite{Fixed19}  Generating Long Sequences with Sparse Transformers
decreases both the computational and memory demands compared to the original ``full'' self-attention\cite{Scale17}, which allows every element to attend to every other prior input. In factorized self-attention, attention is limited to a subset of input elements. Formally, an output matrix is generated using a matrix of input embeddings  $X$ and the connectivity pattern $S=\{S_1,...,S_{n}\}$, where $S_i$ is the set of indices of input vectors attended to by the $i$th output vector.
\eq{
&\text{FacAtt}(X, S) = (A(\mathbf{x}_{i},S_{i}))_{i\in[1,n]}\\
&a(\mathbf{x}_{i}, S_{i}) = \text{softmax}(\frac{(W_{q}\mathbf{x}_{i})K^{T}_{S_{i}}}{\sqrt{d}})\cdot V_{S_{i}} \\
&K_{Si} = (W_{k}\mathbf{x}_{j})_{j\in{S_{i}}} \text{\phantom{abc}} V_{S_i} = (W_{v}\mathbf{x}_{j})_{j\in{S_{i}}} 
}
For full self-attention $S^F_i:=\{j|j\neq i\}$, i.e., all indexes to inputs prior to the $i$th input. In contrast, factorized self-attention has $p$ separate attention heads, where the $m$th head is given by a subset $A_i^{(m)} \subset S^F_i$  with $S_i=A_i^{(m)}$.
For strided self-attention:
\eq{ &A_{i}^{(1)}=\{t,t+1,...i\} \text{ for } t=\max(0,i-l)\\
&A_i^{(2)}=\{j:(i-j)\mod l=0\}}
This pattern is adequate, when structure is similar to the stride-like images. For data without a periodic structure like text, fixed attention can be preferable:
\eq{
&A_{i}^{(1)}=\{j: \lfloor j/l\rfloor = \lfloor i/l \rfloor \}\\
& A_i^{(2)}=\{j: j \mod l \in \{t,t+1,...l\}\}
} where $t=l-c$ with hyperparameter $c$. For instance, with a stride of 128 and $c=8$, all future positions greater than 128 can attend to positions 120-128, all greater 256 to 248-256, and so on.

\smallskip
\noindent\textbf{Multi- and Grouped Query Attention}\cite{shaz19,ain23} 
Multi-query attention(MQA)\cite{shaz19} uses multiple query but only one key head and one value head, which differs from standard muli-head attention. This helps in lowering memory bandwidth for loading keys and values but it reduce quality and training stability. MQA can be added to trained models without MQA with (relatively) little compute \cite{ain23}. Grouped query attention aims at striking a balance between a single and many heads. Per head they use multiple queries rather than just one. This allows to lower quality gaps.

\smallskip
\noindent A \textbf{Residual Attention Network (RAN)}\cite{RAN17} \label{sec:RAN}%\cite{RAN17}  Residual Attention Network for Image Classification
capitalizes on the concept of skip connections and is composed of two branches: a mask branch and a trunk branch. The trunk branch is responsible for processing features and can be any type of network, while the mask branch determines the weights for these features. The output of the attention module is
\eq{H_{i,c}(x)=(1+M_{i,c}(x))\cdot F_{i,c}(X) }
where $i$ is a spatial position and $c$ is a channel. $M(x)$ should be approximatedly 0, $H(x)$ approximates original features $F(x)$.

\smallskip
\noindent\textbf{Large Kernel Attention}\cite{guo22} \label{sec:Visua}%\cite{Visua21}  InferNER: an attentive model leveraging the sentence-level information for Named Entity Recognition in Microblogs
break down a (large-scale) convolution into three more manageable components: a depth-wise dilated convolution, a non-dilated depth-wise convolution, and a 1x1 convolution across channels. Subsequently, an attention map is derived from the outputs of these convolutions.
%not found, this was the closest thing

\smallskip
\noindent\textbf{Sliding Window Attention}\cite{Slidi20} \label{sec:Slidi}%\cite{Slidi20}  Longformer: The Long-Document Transformer
This approach focuses on enhancing the efficiency of attention mechanisms in terms of both time and memory by reducing the number of input pairs considered. Specifically, for a designated window size $w$, each token attends to $\frac{w}{2}$  tokens on either side.

\smallskip
\subsection{Graph Neural Networks }
Graph neural networks, reviwed in \cite{wu2020,kho24}, can be viewed as an extension of CNNs and transformers to accommodate graph-structured data. These networks process information represented as nodes interconnected by edges. We delve into various graph models, focusing on how to derive node embeddings that are applicable to downstream tasks.

%Graph Models
\smallskip
\noindent\textbf{Graph Convolutional Networks (GCN)}\label{sec:GCN}%\cite{GCN16}  Semi-Supervised Classification with Graph Convolutional Networks
\cite{GCN16} use CNNs for semi-supervised learning. They approximate spectral graph convolutions using polynomials of order $k$, which a CNN can compute with $k$ linear layers. This also implies that dependencies are localized, i.e., only up to nodes of distance $k$ from a node.

\smallskip
\noindent\textbf{Scalable Feature Learning for Networks}\label{sec:node2}%\cite{node216}  node2vec: Scalable Feature Learning for Networks
(Node2Vec)\cite{node216} is designed to learn feature vectors that effectively preserve the neighborhood characteristics of nodes within a graph. This method employs random walks to generate samples of neighborhoods, allowing it to capture and represent nodes based on their roles or the communities to which they belong.

\smallskip
\noindent\textbf{Graph Attention Networks}\label{sec:GAT}%\cite{GAT17}  Graph Attention Networks
\cite{GAT17} utilizes masked self-attention layers that enable nodes to adaptively focus on the features of their neighboring nodes. Specifically, node $j$ assigns importance scores to the features of node $i$ based on their connectivity. The use of masking ensures that only connected node pairs are considered. Unlike GCNs, this method allows for different levels of importance to be assigned to nodes within the same neighborhood. Additionally, it avoids the expensive matrix operations associated with eigendecompositions.

%Graph Embeddings
\smallskip
\noindent\textbf{TuckER}\label{sec:TuckE}%\cite{TuckE19}  TuckER: Tensor Factorization for Knowledge Graph Completion
\cite{TuckE19} focuses on using factorization for link prediction within a knowledge graph, where knowledge is structured as (subject, relation, object) triplets. The objective is to predict the existence of a relationship between two entities. This approach models the graph as a binary tensor, with the dimensions representing subjects, relations, and objects. Tucker decompositions are then employed to break down this binary tensor into a core tensor and separate embedding matrices for subjects, relations, and objects.

\smallskip
\noindent\textbf{Embedding by Relational Rotation} \label{sec:Rotat}%\cite{Rotat19}  RotatE: Knowledge Graph Embedding by Relational Rotation in Complex Space
(RotatE)\cite{Rotat19} is used for predicting missing links in knowledge graphs, similar to the previously described TuckER\cite{TuckE19}, but with a focus on modeling additional relational properties like composition and inversion. In this method, entities are embedded in a complex space, and relations are treated as element-wise rotations. These rotations are optimized to effectively transition from one entity to another within the complex embedding space.

\smallskip
\noindent\textbf{Graph Transformer}\label{sec:Graph}%\cite{Graph20}  A Generalization of Transformer Networks to Graphs
\cite{Graph20}   extends the original transformer architecture to handle graph structures by implementing attention mechanisms that consider the connectivity of each node's neighborhood. This adaptation generalizes the concept of position encoding to suit graph data. The model also replaces layer normalization with batch normalization and introduces the capability to learn representations for edges, in addition to node representations.

\section{Discussion}
Our findings suggest that despite many small and creative innovations since the original transformer architecture, there have not been any significant "breakthrough" discoveries that have led to much better leaderboard results. The improvements on models such as large language models within the last few years have been characterized by the enlargement of existing networks such as GPT, the increase of data volume (and quality), focus on computational efficiency, and a shift towards self-supervised learning. This could indicate a need for more daring approaches to research rather than incremental improvements of existing works and it raises the question: 
\emph{Are transformers all we can do?}

Combining different elements or pursuing more radical ideas as outlined in this work could be one way to go beyond transformers. But advancing novel ideas such as state-space models and ideas that have existed for a while such as capsule networks, but have not lived up to their hopes, is not an easy task. We have identified a few general patterns that have been proven effective in many areas and might help in accomplishing this task: 
\begin{itemize}
    \item \textbf{Multi-X} refers to the strategy of deploying the same component repeatedly in parallel configurations, such as employing several residual blocks (ResNeXt) or utilizing multi-head attention mechanisms or multiple models as for mixture of experts. This concept is closely aligned with the principles of "ensemble learning."
    \item \textbf{Higher order layers} involve more complex operations than the linear layers and simple ReLU functions typically used in classical CNNs and MLPs. Examples of these more advanced layers include Mish or attention layers, which facilitate deeper and more nuanced processing. 

    \item \textbf{Data controlled gating} is an instance of a higher order layer. The idea of data controlled gating, e.g., depending on some (possibly transformed) input $X$, another input $X'$ is deemed relevant or ignored, seems powerful. This idea can be found in attention but also in LSTM layers and GELU, where we have ``self-gating''. On a larger scale also mixture of experts perform data controlled gating and, more generally, in transistors. Data controlled gating can also often be interpreted as \textbf{Weighing functions}, which involves the use of parameterized functions to weigh inputs. Instead of simply aggregating inputs, these functions assign weights to them, often derived from learned parameters. These functions act as "gates," allowing the flow of information only within a specific range of input parameters.
    
    \item \textbf{Moving average} refers to the technique of averaging weights, as seen in methods like SGD and BYOL.
    \item \textbf{Decompose}  refers to breaking down matrices into simpler components, as exemplified by methods like TuckER and large kernel attention.
    %\item ``Weighing functions'', i.e., using parameterized weighing functions of inputs can be seen within the attention mechanism but also for GELU units. Therefore, rather than naively aggregating inputs, inputs are weighed and aggregated. The weight might stem from a function with learnt parameters. Such functions can also be seen as ``gates'' that only permit the flow of information within some range of the input parameters.
\end{itemize}

Furthermore, the transformer seems to violate the ``no free lunch theorem'' by Wolpert saying that no model can be best for all (generative) tasks. While fine-tuning models on specific data can yield better (transformer) models for specific tasks, it is not clear, whether specialized architectures might substantially outperform the general transformer in certain sets of tasks, where it seems to outperform. This is analogous to activation functions, where no activation outperforms on all datasets and within all models \cite{dub22}.

Our survey focused on key design elements in building deep learning models. Taking a practical approach, we chose to ignore theoretical works, which should be further explored in future studies. For example, \cite{sieb24} derived a theoretical framework for dynamical systems allowing to better compare model classes such as linear attention and state space models.

Our survey intentionally focused on more recent, yet well-established works, which could be seen as either a strength or a limitation. The selection of papers for the deep learning part was guided by a prominent platform that offers leaderboards. The rise of such platforms, which allow the upload of papers and models and provide information on citations and rankings, offers a fresh perspective compared to traditional survey methods that often choose papers more arbitrarily. Although this approach benefits readers looking for "what works well and what is very promising," it may overlook innovative ideas that need more research to fully demonstrate their potential. This could contribute to the "winner-takes-all" dynamic, reinforcing already successful ideas. Furthermore, such platforms might undergo a hype-cycle, e.g., become very popular for a while before vanishing. Nevertheless, given the vast number of papers, some form of selection is essential for conducting a comprehensive survey of deep learning.

We recognize that online platforms offering leaderboards and related features are highly valuable to the research community and should continue to be developed. However, we found that manual verification was necessary, such as double-checking relevance with Google Scholar citations and reviewing surveys and papers, to identify works and methods that were not accurately listed on the platform.

\smallskip

\section{Conclusions}
We have provided a concise yet thorough overview of the deep learning design landscape, with a focus on transformers and their potential successors. Key works from various significant areas that have emerged in recent years have been summarized. We believe that our holistic overview in a single paper can establish connections that may inspire novel ideas. Additionally, we have identified four patterns that characterize many improvements in this field. To further advance the development of artificial intelligence, it is crucial to generate fundamentally new and successful approaches, as recent improvements, though numerous and often very creative, have primarily been incremental.
%focus on papers not so much categorization of them
%different categorization possible

%\noindent \textbf{Ethical Statement:} There are no ethical issues.

% \section{Declarations}
% \subsection{Availability of data and material }
% All data used is public.

% \subsection{Competing interests}
% The authors declare that they have no  competing interests. 
% \subsection{Ethics Approval}
% Not applicable
% \subsection{Consent to participate}
% Not applicable
% \subsection{Consent for publication}
% Not applicable
 
% \subsection{Funding}
% Not Applicable

% \smallskip

%\bibliographystyle{IEEEtran}

%\bibliographystyle{apalike}
\bibliographystyle{splncs04}
\bibliography{refs}
%{\small \bibliography{refs}}

% Download pages 
% -> summarize higher level mechanisms, usage and stars of Top 10
% (also relative to age)
% Show chart over time
% We state mathematical formulas if can be stated in 2-4 lines

% techniques are difficult often as part of network -> Resnet -> cite if use resnet or if use just res connection
% 0,2,4,6,8
% y=k*x
% sum(y)=k*sum(x)
% k=sum(y)/sum(x)
% k=tot/years

\end{document}